\documentclass[twocolumn]{article}
\usepackage{arxiv}  % ARXIV FORMAT

\usepackage{cite}
\usepackage{amsmath,amssymb,amsfonts}
\usepackage{algorithmic}
\usepackage{graphicx}
\usepackage{textcomp}
\usepackage{xcolor}
\usepackage{fancyhdr}
\usepackage{lipsum} 
\usepackage{booktabs}
\usepackage{array}
\usepackage{multirow}
\usepackage[T1]{fontenc}
\usepackage[inline]{enumitem}

\usepackage{tikz}
\usetikzlibrary{arrows.meta}
\usetikzlibrary{decorations.pathreplacing}
\usetikzlibrary{decorations}
\usepackage{pgfplots}
\pgfplotsset{compat=1.9}
\usepgfplotslibrary{groupplots}
\usetikzlibrary{patterns}

\newcolumntype{M}[1]{>{\centering\arraybackslash}m{#1}}
\newcommand{\No}[0]{N\textsuperscript{\underline{o}}}

\usepackage[pdfencoding=auto, pdfusetitle]{hyperref}
\hypersetup{
  hidelinks,
  urlcolor=red,
  pdftitle={Accommodating Missing Modalities in Time-Continuous Multimodal Emotion Recognition},
  pdfauthor={Juan Vazquez-Rodriguez, Grégoire Lefebvre, Julien Cumin, James L. Crowley},
  pdfkeywords={Affective Computing, Multimodal Emotion Recognition, Machine Learning, Transformers},
}

\title{Accommodating Missing Modalities in Time-Continuous Multimodal Emotion Recognition
}

\author{Juan Vazquez-Rodriguez$^{1,2}$, Grégoire Lefebvre$^{1}$, Julien Cumin$^{1}$, James L. Crowley$^2$\vspace{0.1cm}\\
{$^1$ Orange Innovation, Grenoble, France}\\
{$^2$ Univ. Grenoble Alpes, CNRS, Grenoble INP, LIG, 38000 Grenoble, France}\\
}

\begin{document}

\twocolumn[{%
  \begin{@twocolumnfalse}
    \maketitle
  \end{@twocolumnfalse}
}]

\setcounter{footnote}{0}

\begin{abstract}
Decades of research indicate that emotion recognition is more effective when drawing information from multiple modalities. But what if some modalities are sometimes missing? To address this problem, we propose a novel Transformer-based architecture for recognizing valence and arousal in a time-continuous manner even with missing input modalities. We use a coupling of cross-attention and self-attention mechanisms to emphasize relationships between modalities during time and enhance the learning process on weak salient inputs. Experimental results on the Ulm-TSST dataset show that our model exhibits an improvement of the concordance correlation coefficient evaluation of 37\% when predicting arousal values and 30\% when predicting valence values,  compared to a late-fusion baseline approach.
\end{abstract}


\section{Introduction} \label{section:intro}
Technologies for automatic emotion recognition have been shown valuable for interpersonal communications\cite{zenonosHealthyOfficeMoodRecognition2016}, health and wellness concerns\cite{zenonosHealthyOfficeMoodRecognition2016} and stress management\cite{nunezEffectSocialConnectedness2019}, for example. People express emotions in both verbal and non-verbal manners. Facial expressions, pitch intensity or cardiac rhythms are examples of non-verbal communication.

Using multiple modalities for emotion recognition is advantageous since modalities may be complementary, and should thus improve the performance of the model when used together \cite{ouzarVideoBasedMultimodalSpontaneous2022}. However, in real-world scenarios, there might be cases where a modality might not be available. If for example, the modalities consist of video, audio, and physiological signals, the camera field of view might be obstructed, the microphone can be too far away, or the physiological sensor might be on a wearable device that is not currently worn. Therefore, we need a model capable of handling missing modalities. 

In this work, we extend a multimodal Transformer \cite{gabeurMultimodalTransformerVideo2020} as an encoder to obtain representations from the different modalities and a Transformer decoder \cite{vaswaniAttentionAllYou2017} to process those representations and make predictions. A Transformer-based approach will continue to work in the case of missing modalities, although the performance often decreases \cite{maAreMultimodalTransformers2022}. 
We investigated a learning strategy to improve performance by eliminating the most important modalities during part of the training, so that the model is forced to learn from the less informative ones, that nevertheless may carry valuable features. This has two desirable effects. First, the model improves its performance, by training the model to draw information from all modalities rather than focusing on the most important ones. Second, the model becomes less sensitive to missing modalities, as it learns to handle the case where a modality is not present. 

A critical aspect of multimodal emotion recognition is modeling the complementarity of information from different modalities. In other words, the model should be capable of weighing the different modalities according to their importance. 
We then present a novel approach of using the encoder-decoder attention (cross-attention) of the Transformer decoder to weigh the representations generated by the encoder, making this weighting scheme focus on choosing between modalities rather than paying attention to information from different time-steps. In addition, our Transformer decoder is auto-regressive, meaning that it takes into account past predicted values when doing the current inference, which is important when performing time-continuous predictions.

The main contributions of this research are:
\begin{enumerate*}
    \item we extend a multimodal Transformer-based architecture to perform time-continuous value-continuous multimodal emotion recognition,
    \item we present a novel approach using cross-attention from the Transformer decoder to weigh the importance of different modalities,
    \item and we develop a learning strategy to improve the performance of the model when a modality is missing.
\end{enumerate*}

\section{Related Work} \label{section:related}

\subsection{Time-Continous Multimodal Emotion Recognition} \label{section:timeContEmoRec}
Several works address the problem of time-continuous multimodal emotion recognition. Traditionally, Long Short Term Memory - Recurrent Neural Networks (LSTM-RNN) have been employed to model the temporal relations of the inputs and to consider past predictions when predicting the current time-step \cite{christMuSe2022Multimodal2022a}. Recently, employing Transformer-based approaches \cite{huangMultimodalTransformerFusion2020, zhangContinuousEmotionRecognition2022a, chenTransformerEncoderMultiModal2021} has gained popularity in addressing this task. Some works rely on LSTM-RNNs to complement the attention mechanisms from the Transformer to model the temporal information better \cite{huangMultimodalTransformerFusion2020}, while other works use a pure attention-based approach \cite{zhangContinuousEmotionRecognition2022a, chenTransformerEncoderMultiModal2021}. 

To fuse information from different modalities, some authors use late-fusion combining the outputs of single modality models \cite{liuImprovingDimensionalEmotion2022}, while others employ early-fusion by combining the input features before feeding them into the model \cite{karasTimeContinuousAudiovisualFusion2022}. Some approaches that use Transformer-based techniques have presented more elaborate solutions to aggregate multimodal information. Cross-modal attention \cite{tsaiMultimodalTransformerUnaligned2019} can be used to incorporate information from different modalities \cite{huangMultimodalTransformerFusion2020, karasTimeContinuousAudiovisualFusion2022}. Different from this, Zhang et al. \cite{zhangContinuousEmotionRecognition2022a} group the query vectors from each modality to form a single query vector and do the same for the key and value vectors. Then, they employ the grouped vectors to perform a modified version of the scaled dot-product attention described in the original Transformer paper \cite{vaswaniAttentionAllYou2017}. Chen et al. \cite{chenTransformerEncoderMultiModal2021} and He et al. \cite{heMultimodalTemporalAttention2022} model temporal information using a standard Transformer approach and they model intermodal information through a multimodal attention mechanism.

A disadvantage of these approaches is that, at some point, the features coming from the different modalities are concatenated. This requires that all the modalities need to be present, thus breaking the approach if a modality is missing. Some authors have worked on addressing this situation, and we review some of these works below.

\subsection{Handling Missing Modalities}
There are three main types of approaches to handle missing modalities \cite{zhaoMissingModalityImagination2021}: 
\begin{enumerate*}
    \item learning a joint representation from the different modalities, so only one modality could be used at test time, 
    \item generating the missing modalities from the available ones, and
    \item hiding some modalities during training.
\end{enumerate*}

For the first type, an example is the work of Pham et al. \cite{phamFoundTranslationLearning2019}, where a joint representation is learned by encoding the text into a representation (the joint representation) and generating the other modalities from this representation. At test time, only the text input is needed. For the second type, we have the work of Mittal et al. \cite{mittalM3ERMultiplicativeMultimodal2020}, where the model generates replacement features using a learned linear transformation that converts features from the available modalities into features of the missing one. For the third type, an example is the work of Neverova et al. \cite{neverovaModDropAdaptiveMultiModal2016}, where a carefully designed network is designed so it can still work even with missing modalities. Then, at train time, some modalities are dropped randomly to make the model robust to missing modalities.

Although these approaches make the model robust to missing modalities, a disadvantage of the first type of approach is that it cannot take advantage of using all modalities if they are present at test time.
For the second type of approach, a drawback is that there is no guarantee that the generated representation accurately resembles the missing one.
And to implement the third type of approach, the architecture should be capable of working with missing modalities.
On the contrary, if a Transformer-based approach is used, there is no need to generate the missing modality representations, or do modifications to the architecture so it can work with modalities absent.
In this case, the attention mechanisms simply do not attend to the missing modalities, and it is capable of attending to all of them if they are present.

For the reasons stated in the previous paragraph, many works that use the third type of approach to handle missing modalities use a Transformer-based model. Some examples include the work of Goncalves and Busso \cite{goncalvesAuxFormerRobustApproach2022}, and the work of Parthasarathy and Sundaram \cite{parthasarathyTrainingStrategiesHandle2021}, where they use a cross-modality Transformer to combine audio and visual modalities, improving the robustness of the model to missing modalities by eliminating a modality during training. A disadvantage of using a cross-modality Transformer is that expanding the approach to use more modalities is not straightforward. To overcome this problem, a Multimodal Transformer \cite{gabeurMultimodalTransformerVideo2020} can be employed, like in the work of Ma et al. \cite{maAreMultimodalTransformers2022}, where robustness to missing modalities is increased using a multitasking approach.

The Transformer-based models are well suited to model long and short temporal relations of the inputs and to model the cross-modality dependencies. Nevertheless, in the reviewed Transformer-based approaches, the attention layers have to model the temporal and the intermodal dependencies at the same time. We argue that it can be advantageous to attend only to the cross-modal dependencies when aggregating the multimodal information. In addition, the reviewed approaches do not explicitly consider past predictions when making the current prediction, which we believe is beneficial. 

Our approach is a Transformer-based approach that uses a Multimodal Transformer as encoder, making it suitable for any number of modalities. We also use the novel idea of using the cross-attention from a Transformer decoder \cite{vaswaniAttentionAllYou2017} to weigh the information from the different modalities. The decoder uses only the information of each modality at the current prediction time-step, relieving it from modeling the temporal information, which is done by the encoder. In addition, we explicitly use past predictions to make the current one by employing an auto-regressive approach. To handle missing modalities, we use the approach of hiding some modalities at train time, but different from the state of the art, we employ a technique to find and then hide the important modalities.

\section{Approach} \label{section:approach}
In this section, we provide a detailed explanation of our approach to perform multimodal time-continuous value-continuous emotion recognition. Our objective is to predict values of arousal and valence. We start this section by explaining our encoder that generates multimodal representations. Then we explain our decoder that predicts the values of arousal and valence from those representations. Finally, we describe the loss that we use to train our model.

\subsection{MultiModal Tranformer Encoder (MMTE)} \label{section:trEncoder}

\begin{figure}[tb]
\centering

\begin{tikzpicture}[inner sep=3pt, thick, scale=0.4, every node/.style={scale=0.8}]

\tikzset{
  mynode/.style={execute at begin node=\setlength{\baselineskip}{8pt}}
}

\begin{scope}[shift={(-0.75,9)}]
    \draw[rounded corners=5, fill=yellow!50] (0,0) rectangle (16,2) node[pos=0.5] {\textbf{Transformer Encoder}};
\end{scope}

%Modality 1
\begin{scope}

    % Output features
    \begin{scope}[shift={(0, 11.5)}]
    \node at (0.25, 2) {$r^1_1$};
    \node at (1.75, 2) {$r^1_2$};
    \node at (4.25, 2) {$r^1_T$};
    \node at (3, 0.75) {\textbf{...}};
    
    \draw [fill=red] (0, 0) rectangle (0.5, 1.5);
    \draw [fill=red] (1.5, 0) rectangle (2, 1.5);
    \draw [fill=red] (4, 0) rectangle (4.5, 1.5);

    \draw [semithick, {Latex[length=1mm]}-] (0.25, 0) -- +(0, -0.5);
    \draw [semithick, {Latex[length=1mm]}-] (1.75, 0) -- +(0, -0.5);
    \draw [semithick, {Latex[length=1mm]}-] (4.25, 0) -- +(0, -0.5);
    \end{scope}

    % Transformer inputs
    \begin{scope}[shift={(-0.25, 7.5)}]
    \draw [semithick] (0.5, 0) -- +(0, 0.25);
    \draw [semithick] (2.0, 0) -- +(0, 0.25);
    \draw [semithick] (4.5, 0) -- +(0, 0.25);

    \node at (0.6, 0.6) {\footnotesize $f^1_1$};
    \node at (2.1, 0.6) {\footnotesize $f^1_2$};
    \node at (4.6, 0.6) {\footnotesize $f^1_T$};        

    \draw [semithick, -{Latex[length=1mm]}] (0.5, 1) -- +(0, 0.5);
    \draw [semithick, -{Latex[length=1mm]}] (2.0, 1) -- +(0, 0.5);
    \draw [semithick, -{Latex[length=1mm]}] (4.5, 1) -- +(0, 0.5);
    \end{scope}

    % Modality Encoding
    \begin{scope}[shift={(-0.25,6.5)}]
    \draw [fill=red!50] (0,0) rectangle (1,1) node[pos=0.5] {$e^1$};
    \draw [fill=red!50] (1.5,0) rectangle (2.5,1) node[pos=0.5] {$e^1$};
    \node at (3.25, 0.5) {\textbf{...}};
    \draw [fill=red!50] (4,0) rectangle (5,1) node[pos=0.5] {$e^1$};
    \end{scope}
    
    % Addition 2
    \begin{scope}[shift={(-0.25, 6)}]
    \draw [semithick] (0.5, 0.25) -- +(0, 0.25);
    \draw [semithick] (2, 0.25) -- +(0, 0.25);
    \draw [semithick] (4.5, 0.25) -- +(0, 0.25);
    
    \draw [semithick] (0.5,0) circle [radius=0.25] node {+};
    \draw [semithick] (2,0) circle [radius=0.25] node {+};
    \draw [semithick] (4.5,0) circle [radius=0.25] node {+};
    
    \draw [semithick] (0.5, -0.25) -- +(0, -0.25);
    \draw [semithick] (2, -0.25) -- +(0, -0.25);
    \draw [semithick] (4.5, -0.25) -- +(0, -0.25);
    \end{scope}
    
    % Positional Encoding
    \begin{scope}[shift={(-0.25,4.5)}]
    \draw [fill=black!5] (0,0) rectangle (1,1) node[pos=0.5] {$p_1$};
    \draw [fill=black!20] (1.5,0) rectangle (2.5,1) node[pos=0.5] {$p_2$};
    \node at (3.25, 0.5) {\textbf{...}};
    \draw [fill=black!40] (4,0) rectangle (5,1) node[pos=0.5] {$p_T$};
    \end{scope}
    
    % Addition 1
    \begin{scope}[shift={(-0.25,4)}]
    \draw [semithick] (0.5, 0.25) -- +(0, 0.25);
    \draw [semithick] (2, 0.25) -- +(0, 0.25);
    \draw [semithick] (4.5, 0.25) -- +(0, 0.25);
    
    \draw [semithick] (0.5,0) circle [radius=0.25] node {+};
    \draw [semithick] (2,0) circle [radius=0.25] node {+};
    \draw [semithick] (4.5,0) circle [radius=0.25] node {+};
    
    \draw [semithick] (0.5, -0.25) -- +(0, -0.25);
    \draw [semithick] (2, -0.25) -- +(0, -0.25);
    \draw [semithick] (4.5, -0.25) -- +(0, -0.25);

    \end{scope}
    
    % TCN
    \begin{scope}[shift={(-0.75,1.75)}]
    \draw [semithick] (1.0, 1) -- +(0, 0.25);
    \draw [semithick] (2.5, 1) -- +(0, 0.25);
    \draw [semithick] (5.0, 1) -- +(0, 0.25);

    \draw [black, fill=orange!30] (0, 0) -- +(6, 0) -- +(5.5, 1) -- +(0.5, 1) -- cycle node at(3, 0.5) {\textbf{TCN\textsuperscript{1}}};

    \node at (1.2, 1.5) {\footnotesize $a^1_1$};
    \node at (2.7, 1.5) {\footnotesize $a^1_2$};
    \node at (5.2, 1.5) {\footnotesize $a^1_T$};
    \end{scope}
    
    % Input features
    \begin{scope}[shift={(0,-0.25)}]
    \draw [semithick, -{Latex[length=1mm]}] (0.25, 1.5) -- +(0, 0.5);
    \draw [semithick, -{Latex[length=1mm]}] (1.75, 1.5) -- +(0, 0.5);
    \draw [semithick, -{Latex[length=1mm]}] (4.25, 1.5) -- +(0, 0.5);
    
    \draw [fill=red] (0.0, 0) rectangle (0.5, 1.5);
    \draw [fill=red] (1.5, 0) rectangle (2.0, 1.5);
    \draw [fill=red] (4.0, 0) rectangle (4.5, 1.5);
    
    \node at (0.25, -0.5) {$x^1_1$};
    \node at (1.75, -0.5) {$x^1_2$};
    \node at (4.25, -0.5) {$x^1_T$};
    
    \node at (3, 0.75) {\textbf{...}};
    \end{scope}
        
\end{scope}

% dots
\begin{scope}[shift={(7.5,0)}]
    \node at (0, 0.75) {\Large \textbf{...}};
    \node at (0, 2.25) {\Large \textbf{...}};
    \node at (0, 5) {\Large \textbf{...}};
    \node at (0, 7) {\Large \textbf{...}};
    \node at (0, 11.75) {\Large \textbf{...}};
\end{scope}

% Modality M
\begin{scope}[shift={(10,0)}]

    % Output features
    \begin{scope}[shift={(0, 11.5)}]
    \node at (0.25, 2) {$r^M_1$};
    \node at (1.75, 2) {$r^M_2$};
    \node at (4.25, 2) {$r^M_T$};
    \node at (3, 0.75) {\textbf{...}};
    
    \draw [fill=blue] (0, 0) rectangle (0.5, 1.5);
    \draw [fill=blue] (1.5, 0) rectangle (2, 1.5);
    \draw [fill=blue] (4, 0) rectangle (4.5, 1.5);

    \draw [semithick, {Latex[length=1mm]}-] (0.25, 0) -- +(0, -0.5);
    \draw [semithick, {Latex[length=1mm]}-] (1.75, 0) -- +(0, -0.5);
    \draw [semithick, {Latex[length=1mm]}-] (4.25, 0) -- +(0, -0.5);

    \node[mynode, right, align=left] at (5.5, 0.75) {\footnotesize Encoder \\ \footnotesize Outputs};  
    \end{scope}

    % Transformer inputs
    \begin{scope}[shift={(-0.25, 7.5)}]
    \draw [semithick] (0.5, 0) -- +(0, 0.25);
    \draw [semithick] (2.0, 0) -- +(0, 0.25);
    \draw [semithick] (4.5, 0) -- +(0, 0.25);

    \node at (0.7, 0.6) {\footnotesize $f^M_1$};
    \node at (2.2, 0.6) {\footnotesize $f^M_2$};
    \node at (4.7, 0.6) {\footnotesize $f^M_T$};        

    \draw [semithick, -{Latex[length=1mm]}] (0.5, 1) -- +(0, 0.5);
    \draw [semithick, -{Latex[length=1mm]}] (2.0, 1) -- +(0, 0.5);
    \draw [semithick, -{Latex[length=1mm]}] (4.5, 1) -- +(0, 0.5);
    \end{scope}

    % Modality Encoding
    \begin{scope}[shift={(-0.25,6.5)}]
    \draw [fill=blue!30] (0,0) rectangle (1,1) node[pos=0.5] {$e^M$};
    \draw [fill=blue!30] (1.5,0) rectangle (2.5,1) node[pos=0.5] {$e^M$};
    \draw [fill=blue!30] (4,0) rectangle (5,1) node[pos=0.5] {$e^M$};

    \node at (3.25, 0.5) {\textbf{...}};
    \node[mynode, right, align=left] at (5.75, 0.5) {\footnotesize Modality \\ \footnotesize Encodings};  
    \end{scope}
    
    % Addition 2
    \begin{scope}[shift={(-0.25, 6)}]
    \draw [semithick] (0.5, 0.25) -- +(0, 0.25);
    \draw [semithick] (2, 0.25) -- +(0, 0.25);
    \draw [semithick] (4.5, 0.25) -- +(0, 0.25);
    
    \draw [semithick] (0.5,0) circle [radius=0.25] node {+};
    \draw [semithick] (2,0) circle [radius=0.25] node {+};
    \draw [semithick] (4.5,0) circle [radius=0.25] node {+};
    
    \draw [semithick] (0.5, -0.25) -- +(0, -0.25);
    \draw [semithick] (2, -0.25) -- +(0, -0.25);
    \draw [semithick] (4.5, -0.25) -- +(0, -0.25);
    \end{scope}
    
    % Positional Encoding
    \begin{scope}[shift={(-0.25,4.5)}]
    \draw [fill=black!5] (0,0) rectangle (1,1) node[pos=0.5] {$p_1$};
    \draw [fill=black!20] (1.5,0) rectangle (2.5,1) node[pos=0.5] {$p_2$};
    \draw [fill=black!40] (4,0) rectangle (5,1) node[pos=0.5] {$p_T$};

    \node at (3.25, 0.5) {\textbf{...}};
    \node[mynode, right, align=left] at (5.75, 0.5) {\footnotesize Positional \\ \footnotesize Encodings};  
    \end{scope}
    
    % Addition 1
    \begin{scope}[shift={(-0.25,4)}]
    \draw [semithick] (0.5, 0.25) -- +(0, 0.25);
    \draw [semithick] (2, 0.25) -- +(0, 0.25);
    \draw [semithick] (4.5, 0.25) -- +(0, 0.25);
    
    \draw [semithick] (0.5,0) circle [radius=0.25] node {+};
    \draw [semithick] (2,0) circle [radius=0.25] node {+};
    \draw [semithick] (4.5,0) circle [radius=0.25] node {+};
    
    \draw [semithick] (0.5, -0.25) -- +(0, -0.25);
    \draw [semithick] (2, -0.25) -- +(0, -0.25);
    \draw [semithick] (4.5, -0.25) -- +(0, -0.25);

    \end{scope}
    
    % TCN
    \begin{scope}[shift={(-0.75,1.75)}]
    \draw [semithick] (1.0, 1) -- +(0, 0.25);
    \draw [semithick] (2.5, 1) -- +(0, 0.25);
    \draw [semithick] (5.0, 1) -- +(0, 0.25);

    \draw [black, fill=orange!30] (0, 0) -- +(6, 0) -- +(5.5, 1) -- +(0.5, 1) -- cycle node at(3, 0.5) {\textbf{TCN\textsuperscript{M}}};

    \node at (1.2, 1.5) {\footnotesize $a^M_1$};
    \node at (2.7, 1.5) {\footnotesize $a^M_2$};
    \node at (5.2, 1.5) {\footnotesize $a^M_T$};
    \end{scope}
    
    % Input features
    \begin{scope}[shift={(0,-0.25)}]
    \draw [semithick, -{Latex[length=1mm]}] (0.25, 1.5) -- +(0, 0.5);
    \draw [semithick, -{Latex[length=1mm]}] (1.75, 1.5) -- +(0, 0.5);
    \draw [semithick, -{Latex[length=1mm]}] (4.25, 1.5) -- +(0, 0.5);
    
    \draw [fill=blue] (0, 0) rectangle (0.5, 1.5);
    \draw [fill=blue] (1.5, 0) rectangle (2, 1.5);
    \draw [fill=blue] (4, 0) rectangle (4.5, 1.5);
    
    \node at (0.25, -0.5) {$x^M_1$};
    \node at (1.75, -0.5) {$x^M_2$};
    \node at (4.25, -0.5) {$x^M_T$};
    
    \node at (3, 0.75) {\textbf{...}};
    \node[mynode, right, align=left] at (5.5, 0.75) {\footnotesize Input \\ \footnotesize Features};  
    \end{scope}
        
\end{scope}

\end{tikzpicture}

\caption{MultiModal Transformer Encoder (MMTE).}
\label{fig:trEncoder}
\end{figure}
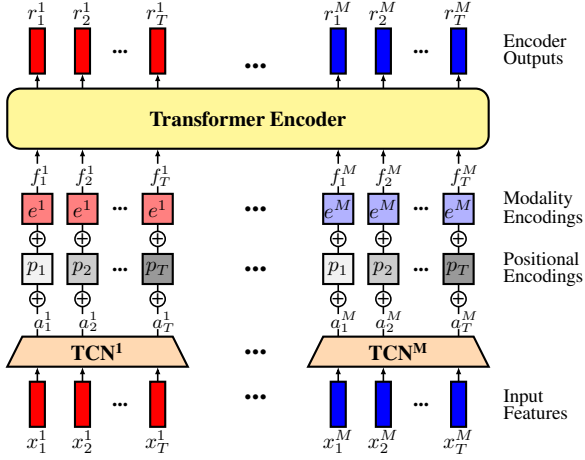

We depict our MultiModal Transformer Encoder (MMTE) in Figure \ref{fig:trEncoder}. Our MMTE is based on the work by Gabeur et al. \cite{gabeurMultimodalTransformerVideo2020}. The inputs for our encoder are features extracted from raw data from the different modalities. We discuss the features we use in Section \ref{section:experiments}. 

The first step in our MMTE architecture is to process each modality individually using a Temporal Convolutional Network (TCN) \cite{baiEmpiricalEvaluationGeneric2018} to model local temporal information, similarly to \cite{chenTransformerEncoderMultiModal2021}. Our model learns a different TCN for each modality. We define $x^m_t \in \mathbb{R}^{d_\text{modality}}$ as the feature corresponding to modality $m$ at time-step $t$. If we denote $[x^m_1, \dots, x^m_T]$ as the sequence with length $T$ of features corresponding to modality $m$, then during this step we have:
\begin{equation} \label{eq:tcn}
 [a^m_1, \dots, a^m_T] = \text{TCN}^m([x^m_1, \dots, x^m_T]),
\end{equation}
where $a^m_t \in \mathbb{R}^{d_\text{model}}$. For all modalities, the TCN output will have a common size $d_\text{model}$. 

The next step is to add positional encodings that allow the Transformer to take into account the actual order of the sequence \cite{vaswaniAttentionAllYou2017}.
If the sequence of positional encodings is $P = [p_1, \dots, p_T]$, with $p_t \in \mathbb{R}^{d_\text{model}}$, then the output of this step is
\begin{equation} \label{eq:posEncoding}
 [a^m_1 + p_1, \dots, a^m_T + p_T].
\end{equation}
The elements of $P$ are parameters that are learned during the training of the whole architecture.

The Transformer also needs to differentiate each modality to process cross-modality information. To do this, we follow the original multimodal Transformer from Gabeur et al., and add modality encodings. Similar to positional encodings, these modality encodings are learned during training. For each modality $m$, an encoding $e^m \in \mathbb{R}^{d_\text{model}}$ is added to the input. The output of this step is then
\begin{equation} \label{eq:modEncoding}
 [a^m_1 + p_1 + e^m, \dots, a^m_T + p_T + e^m].
\end{equation}

We then concatenate the sequences from all modalities to have a single sequence. If we define the input corresponding to modality $m$ at time-step $t$ as $f^m_t = a^m_t + p_t, + e^m$, and if we have $M$ modalities, the concatenated input sequence is then
\begin{equation} \label{eq:inputSeq}
\begin{split}
 [f^1_1, \dots, f^1_T, \dots, f^M_1, \dots, f^M_T].
\end{split}
\end{equation}
We process this sequence using a Transformer encoder. The output representations $r^m_t$ of the Transformer are given by
\begin{equation} \label{eq:transformerEncoder}
 \begin{split}
 [r^1_1, \dots, r^1_T, \dots, r^M_1, &\dots, r^M_T] = \\
 \text{Transformer Encoder}&([f^1_1, \dots, f^1_T, \dots, f^M_1, \dots, f^M_T]).
\end{split}
\end{equation}

Following \cite{chenTransformerEncoderMultiModal2021}, we employ a bidirectional attention mask in the Transformer encoder. When processing a specific time-step, this mask \textit{hides} the inputs that are farther than \textit{mask\_length} positions in the future and in the past. This allows the model to concentrate on recent information, and not to worry about information too far in time that probably does not influence the current emotional state. Note that we are not hiding complete modalities, therefore this technique is not intended to make the model robust to missing modalities.

\subsection{Auto-regressive MultiModal Transformer Decoder (AMMTD)} \label{section:trDecoder}

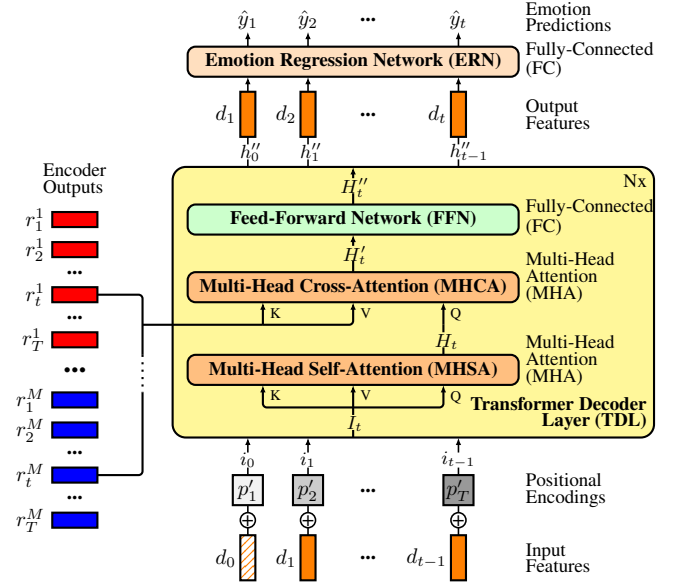
\begin{figure}[tb]
\centering

\begin{tikzpicture}[inner sep=3pt, thick, scale=0.4, every node/.style={scale=0.8},
    mynode/.style={execute at begin node=\setlength{\baselineskip}{8pt}}]

% Encoder Outputs
\begin{scope}[shift={(0,12.25)}]

    \node[mynode, above, align=left] at (0.75, 0.25) {\footnotesize Encoder \\ \footnotesize Outputs};

    % Ouputs mod 1
    \begin{scope}
    \draw [fill=red] (0,  0.0) rectangle (1.5, -0.5);
    \draw [fill=red] (0, -1.0) rectangle (1.5, -1.5);
    \draw [fill=red] (0, -2.5) rectangle (1.5, -3.0);
    \draw [fill=red] (0, -4.0) rectangle (1.5, -4.5);

    \node [left] at (0, -0.25) {$r^1_1$};
    \node [left] at (0, -1.25) {$r^1_2$};
    \node [left] at (0, -2.75) {$r^1_t$};
    \node [left] at (0, -4.25) {$r^1_T$};

    \node at (0.75, -2.0) {\textbf{...}};    
    \node at (0.75, -3.5) {\textbf{...}};    
    \end{scope}

    \node at (0.75, -5.25) {\Large \textbf{...}};   

    % Ouputs mod 2
    \begin{scope}[shift={(0,-6)}]
    \draw [fill=blue] (0,  0.0) rectangle (1.5, -0.5);
    \draw [fill=blue] (0, -1.0) rectangle (1.5, -1.5);
    \draw [fill=blue] (0, -2.5) rectangle (1.5, -3.0);
    \draw [fill=blue] (0, -4.0) rectangle (1.5, -4.5);

    \node [left] at (0, -0.25) {$r^M_1$};
    \node [left] at (0, -1.25) {$r^M_2$};
    \node [left] at (0, -2.75) {$r^M_t$};
    \node [left] at (0, -4.25) {$r^M_T$};

    \node at (0.75, -2.0) {\textbf{...}};    
    \node at (0.75, -3.5) {\textbf{...}};    
    \end{scope}
    
\end{scope}

% AMMTD
\begin{scope}[shift={(4,0)}]

    % ERN
    \begin{scope}[shift={(0.5, 16.75)}]
    \draw [semithick, -{Latex[length=1mm]}] (2, 1) -- +(0, 0.5);
    \draw [semithick, -{Latex[length=1mm]}] (4, 1) -- +(0, 0.5);
    \draw [semithick, -{Latex[length=1mm]}] (9, 1) -- +(0, 0.5);
    \node at (6, 1.9) {\textbf{...}};

    \node at (2, 1.9) {$\hat{y}_1$};
    \node at (4, 1.9) {$\hat{y}_2$};
    \node at (9, 1.9) {$\hat{y}_t$};    
    \node[mynode, right, align=left] at (11, 2) {\footnotesize Emotion \\ \footnotesize Predictions};

    \draw[rounded corners=3, fill=orange!25] (0,0) rectangle (11,1) node [pos=0.5]{\small \textbf{Emotion Regression Network (ERN)}};
    \node[mynode, above right, align=left] at (11, -0.25) {\footnotesize Fully-Connected \\ \footnotesize (FC)};
    \end{scope}

    % Output features
    \begin{scope}[shift={(2.25, 14.75)}]
    \draw [semithick, -{Latex[length=1mm]}] (0.25, 1.5) -- +(0, 0.5);
    \draw [semithick, -{Latex[length=1mm]}] (2.25, 1.5) -- +(0, 0.5);
    \draw [semithick, -{Latex[length=1mm]}] (7.25, 1.5) -- +(0, 0.5);

    \draw [fill=orange] (0, 0) rectangle (0.5, 1.5);
    \draw [fill=orange] (2, 0) rectangle (2.5, 1.5);
    \draw [fill=orange] (7, 0) rectangle (7.5, 1.5);

    \node [left] at (0.1, 0.75) {$d_1$};
    \node [left] at (2.1, 0.75) {$d_2$};
    \node [left] at (7.1, 0.75) {$d_t$};    
    \node at (4.25, 0.75) {\textbf{...}};

    \node[mynode, right, align=left] at (9.25, 0.75) {\footnotesize Output \\ \footnotesize Features};  
    \end{scope}

    % TDL outputs
    \begin{scope}[shift={(2.5, 13.75)}]
    \draw [semithick] (0, 0) -- +(0, 0.25);
    \draw [semithick] (2, 0) -- +(0, 0.25);
    \draw [semithick] (7, 0) -- +(0, 0.25);

    \node at (0.1, 0.5) {\footnotesize $h''_0$};
    \node at (2.1, 0.5) {\footnotesize $h''_1$};
    \node at (7.35, 0.5) {\footnotesize $h''_{t-1}$};

    \draw [semithick] (0, 0.75) -- +(0, 0.25);
    \draw [semithick] (2, 0.75) -- +(0, 0.25);
    \draw [semithick] (7, 0.75) -- +(0, 0.25);
    \end{scope}

    % Transformer Decoder
    \begin{scope}[shift={(0,4.75)}]
        \draw[rounded corners=5, fill=yellow!50] (0,0) rectangle (16,9);
        \node[mynode, above left, align=right] at (16, 0) {\footnotesize \textbf{Transformer} \textbf{Decoder} \\ \footnotesize \textbf{Layer (TDL)}};  
        \node[below left] at (16,9) {\small Nx};
        
        % Out
        \begin{scope}[shift={(6, 7.75)}]
        \draw (0,0) -- (0, 0.375);
        \draw[-{Latex[length=1mm]}] (0,0.875) -- (0,1.25);
        \node at (0.05, 0.57) {\footnotesize $H''_t$};
        \end{scope}

        % FFN
        \begin{scope}[shift={(0.5, 6.75)}]
        \draw[rounded corners=3, fill=green!20] (0,0) rectangle (11,1) node [pos=0.5]{\small \textbf{Feed-Forward Network (FFN)}};
        \node[mynode, above right, align=left] at (11, -0.25) {\footnotesize Fully-Connected \\ \footnotesize (FC)};
        \end{scope}

        % Connection MHCA - FFN
        \begin{scope}[shift={(6, 5.5)}]
        \draw (0,0) -- (0, 0.375);
        \draw[-{Latex[length=1mm]}] (0,0.875) -- (0,1.25);
        \node at (0.05, 0.57) {\footnotesize $H'_t$};
        \end{scope}

        % MHCA
        \begin{scope}[shift={(0.5, 4.5)}]
        \draw[rounded corners=3, fill=orange!50] (0,0) rectangle (11,1) node [pos=0.5]{\small \textbf{Multi-Head Cross-Attention (MHCA)}};
        \node[mynode, above right, align=left] at (11, -0.25) {\footnotesize Multi-Head \\ \footnotesize Attention \\ \footnotesize (MHA) };
        \end{scope}

        % Connection MHSA - MHCA
        \begin{scope}[shift={(6, 2.75)}]
        \draw (3,0) -- (3,0.25);
        \draw (3,0.75) -- (3,1);
        \node at (3.1, 0.45) {\footnotesize $H_t$};
        
        \draw[rounded corners=1.5, -{Latex[length=1mm]}] (3,1) -- (3,1.75);
        \draw[rounded corners=1.5, -{Latex[length=1mm]}] (-3,1) -- (0, 1) -- (0, 1.75);
        \draw[{Latex[length=1mm]}-] (-3,1.75) -- (-3,1);
        \node[below right] at (-3, 1.80) {\scriptsize K};
        \node[below right] at (0, 1.80) {\scriptsize V};
        \node[below right] at (3, 1.80) {\scriptsize Q};
        \end{scope}

        % MHSA
        \begin{scope}[shift={(0.5, 1.75)}]
        \draw[rounded corners=3, fill=orange!50] (0,0) rectangle (11,1) node [pos=0.5]{\small \textbf{Multi-Head Self-Attention (MHSA)}};
        \node[mynode, above right, align=left] at (11, -0.25) {\footnotesize Multi-Head \\ \footnotesize Attention \\ \footnotesize (MHA) };
        \end{scope}

        % Bottom input
        \begin{scope}[shift={(6, 0)}]
        \draw (0,0) -- (0,0.25);
        \draw (0,0.75) -- (0,1);
        \node at (0.05, 0.45) {\footnotesize $I_t$};
        
        \draw[rounded corners=1.5, -{Latex[length=1mm]}] (0,1) -- (0,1.75);
        \draw[rounded corners=1.5, {Latex[length=1mm]}-{Latex[length=1mm]}] (-3,1.75) -- (-3,1) -- (3, 1) -- (3, 1.75);
        \node[below right] at (-3, 1.80) {\scriptsize K};
        \node[below right] at (0, 1.80) {\scriptsize V};
        \node[below right] at (3, 1.80) {\scriptsize Q};
        \end{scope}

    \end{scope}

    % Transformer inputs
    \begin{scope}[shift={(2.5, 3.5)}]
    \draw [semithick] (0, 0) -- +(0, 0.25);
    \draw [semithick] (2, 0) -- +(0, 0.25);
    \draw [semithick] (7, 0) -- +(0, 0.25);

    \node at (0, 0.5) {\footnotesize $i_0$};
    \node at (2, 0.5) {\footnotesize $i_1$};
    \node at (7, 0.5) {\footnotesize $i_{t-1}$};

    \draw [semithick, -{Latex[length=1mm]}] (0, 0.75) -- +(0, 0.5);
    \draw [semithick, -{Latex[length=1mm]}] (2, 0.75) -- +(0, 0.5);
    \draw [semithick, -{Latex[length=1mm]}] (7, 0.75) -- +(0, 0.5);
    \end{scope}
    
    % Positional Encoding
    \begin{scope}[shift={(2,2.5)}]
    \draw [fill=black!5] (0,0) rectangle (1,1) node[pos=0.5] {$p'_1$};
    \draw [fill=black!20] (2,0) rectangle (3,1) node[pos=0.5] {$p'_2$};
    \draw [fill=black!40] (7,0) rectangle (8,1) node[pos=0.5] {$p'_T$};

    \node at (4.5, 0.5) {\textbf{...}};
    \node[mynode, right, align=left] at (9.5, 0.5) {\footnotesize Positional \\ \footnotesize Encodings};  
    \end{scope}
    
    % Addition 1
    \begin{scope}[shift={(2.5,2)}]
    \draw [semithick] (0, 0.25) -- +(0, 0.25);
    \draw [semithick] (2, 0.25) -- +(0, 0.25);
    \draw [semithick] (7, 0.25) -- +(0, 0.25);
    
    \draw [semithick] (0,0) circle [radius=0.25] node {+};
    \draw [semithick] (2,0) circle [radius=0.25] node {+};
    \draw [semithick] (7,0) circle [radius=0.25] node {+};
    
    \draw [semithick] (0, -0.25) -- +(0, -0.25);
    \draw [semithick] (2, -0.25) -- +(0, -0.25);
    \draw [semithick] (7, -0.25) -- +(0, -0.25);
    \end{scope}
        
    % Input features
    \begin{scope}[shift={(2.25,0)}]    
    \draw [pattern=north east lines, pattern color=orange] (0, 0) rectangle (0.5, 1.5);
    \draw [fill=orange] (2, 0) rectangle (2.5, 1.5);
    \draw [fill=orange] (7, 0) rectangle (7.5, 1.5);
    
    \node [left] at (0.1, 0.75) {$d_0$};
    \node [left] at (2.1, 0.75) {$d_1$};
    \node [left] at (7.1, 0.75) {$d_{t-1}$};
    
    \node at (4.25, 0.75) {\textbf{...}};
    \node[mynode, right, align=left] at (9.25, 0.75) {\footnotesize Input \\ \footnotesize Features};  
    \end{scope}
        
\end{scope}

%Cross attention
\begin{scope}
        \draw[rounded corners=1.5] (1.5,9.5) -- (3,9.5) -- (3, 7.5);
        \draw[dotted] (3, 7.5) -- (3, 6.25);
        \draw[rounded corners=1.5] (1.5,3.5) -- (3,3.5) -- (3, 6.25);
        \draw(3,8.5) -- (7,8.5);
\end{scope}

\end{tikzpicture}

\caption{Auto-regressive MultiModal Transformer Decoder (AMMTD). Decoder when predicting the emotion value at time $t$. $Nx$ means that there are $N$ stacked TDL layers.}
\label{fig:trDecoder}
\end{figure}

One of the contributions of this paper is to develop a decoder that predicts emotion from the multimodal representations given by the encoder. To do this, we design an Auto-regressive MultiModal Transformer Decoder (AMMTD). This decoder has two important characteristics: first, it takes previous predictions into account to determine the current emotion; second, it aggregates the representations of the different modalities, giving more weight to the more important ones.

A Transformer Decoder Layer (TDL) \cite{vaswaniAttentionAllYou2017} is composed of a Multi-Head Self-Attention module (MHSA), followed by a Multi-Head Cross-Attention module (MHCA), and followed by a fully-connected Feed-Forward Network (FFN). Residual connections are used around each of these three components. The Multi-Head Attention (MHA) mechanism in MHSA and MHCA projects a query vector $q$ from a given position to a key vector $k$ from another position to determine the attention (i.e. the weight) given to a value $v$ associated with the position of $k$. The final value is the weighted sum of the $v$ vectors from the different positions. We denote the MHA mechanism as 
\begin{equation} \label{eq:mha}
\text{MHA}(Q, K, V),
\end{equation}
where the three parameters $Q$, $K$, and $V$ indicate the sequence used as query, key, and value, respectively. More details about MHA can be found in the original Transformer paper \cite{vaswaniAttentionAllYou2017}. 

Our AMMTD architecture is depicted in Figure \ref{fig:trDecoder}. It is composed of a stack of TDL followed by an Emotion Regression Network (ERN). The MHSA module in the TDL uses self-attention to attend to previous predictions. To do this, we use auto-regression, meaning that the previously generated outputs are used as inputs to the decoder. Note that we cannot use the output of the ERN, i.e. the predicted emotion values $\hat{y}$, since we are predicting continuous outputs. Instead, we use the features generated by the top TDL. When predicting the emotion value at time-step $t$, the TDL stack should have generated a sequence $[d_1, \dots, d_{t-1}]$ with $d_i \in \mathbb{R}^{d_\text{model}}$. Then, the decoder input is
\begin{equation} \label{eq:decInput}
 [d_0, d_1, \dots, d_{t-1}],
\end{equation}
where $d_0 \in \mathbb{R}^{d_\text{model}}$ is a randomly initialized vector.

As for our encoder, we learn positional encodings  $p'_t \in \mathbb{R}^{d_\text{model}}$ for our decoder and add them to the inputs before feeding them to the TDL stack. Thus, when performing the prediction at time-step $t$, the input sequence $I_t = [i_0, i_1, \dots, i_{t-1}]$ with $I_t \in \mathbb{R}^{t \times d_\text{model}}$ becomes
\begin{equation} \label{eq:posEncodingDec}
I_t = [d_0 + p'_0, d_1 + p'_1, \dots, d_{t-1} + p'_{t-1}].
\end{equation}

Inside the TDL, the features are first processed by the MHSA module. This module uses self-attention to integrate information from its own inputs. This means that the query, key, and value for the MHSA all come from the input sequence. To preserve the auto-regressive property, we make sure that a given input at a certain time-step can only attend inputs from past time-steps. Using Expression \ref{eq:mha}, the sequence of features $H_t = [h_0, h_1, \dots, h_{t-1}]$ with $H_t \in \mathbb{R}^{t \times d_\text{model}}$ at the output of the MHSA module is
\begin{equation} \label{eq:mhsa}
 H_t = \text{MHA}(I_t, I_t, I_t).
\end{equation}

The sequence of features $H_t$ is then processed with the MHCA module, which is used to incorporate information from the input modalities. Specifically, the MHCA module attends to the outputs of the encoder. This means that the query comes from the output sequence of the MHSA, and the key and value come from the output of the encoder. If we are predicting the emotion value at time-step $t$, the MHCA attends only to the outputs of each modality corresponding to this time step. The output sequence of the MHCA, $H'_t \in \mathbb{R}^{t \times d_\text{model}}$, using the output of the encoder from Equation \ref{eq:transformerEncoder} and the output of the MHSA from Equation \ref{eq:mhsa}, is
\begin{equation} \label{eq:mhca}
 H'_t = \text{MHA}([h_0, h_1, \dots, h_{t-1}], [r^1_t, \dots, r^M_t], [r^1_t, \dots, r^M_t]),
\end{equation}

Note that we force the model to only attend to the encoder outputs at time $t$ instead of attending to all encoder outputs (or other outputs around $t$) because we want that the MHCA focuses only on finding the best weighting between the different modalities. We want to avoid the MHCA having to weigh which other time-steps in the different modalities might be important. Moreover, this restricts the information flow between modalities, which has been demonstrated to be beneficial \cite{nagraniAttentionBottlenecksMultimodal2021}, because it forces the shared representation to condense the most significant information.

The final step in the TDL is processing each feature of the sequence $H'_t$ through a fully connected feed-forward network (FFN), applied independently to each position:
\begin{equation} \label{eq:ffn}
H''_t = \text{FFN}(H'_t).
\end{equation}

If the TDL stack has more than one layer, the sequence from Equation \ref{eq:ffn} becomes the input of the next layer. Concretely, the new layer implements Equations \ref{eq:mhsa}, \ref{eq:mhca}, and \ref{eq:ffn} using as input $I_t = H''_t$.

For the last TLD layer, the sequence in Equation \ref{eq:ffn} is the newly generated sequence $[d_1, \dots, d_t]$ that will be used as input for the decoder to predict the emotion value for the next time-step, thus if $H''_t = [h''_0, h''_1, \dots, h''_{t-1}]$, then $d_i = h''_{i-1}$ with $i \in [1, t]$.

Once the complete output sequence $D = [d_1, \dots, d_T]$ has been generated, the final step is to process it with the Emotion Regression Network (ERN). As shown in Figure \ref{fig:trDecoder}, our ERN is a Fully Connected (FC) layer that independently processes each element of the sequence $D$ to predict the emotion values for each time-step, producing the predicted sequence $[\hat{y}_1, \dots, \hat{y}_T]$.

\subsection{Loss Function}
As suggested in previous works that address the problem of recognizing emotion in a time-continuous manner \cite{christMuSe2022Multimodal2022a, heMultimodalTemporalAttention2022, liuImprovingDimensionalEmotion2022}, we use the concordance correlation coefficient (CCC) \cite{linConcordanceCorrelationCoefficient1989} as the loss to train our model. Specifically, the loss is
\begin{equation} \label{eq:loss}
\begin{split}
 \mathcal{L} &= 1 - \text{CCC} \\
 \text{CCC} &= \frac{2\rho\sigma_{\hat{y}}\sigma_y}{\sigma^2_{\hat{y}} + \sigma^2_y + (\mu_{\hat{y}} - \mu_y)^2},
\end{split}
\end{equation}
where $\rho$ is the Pearson correlation coefficient between the predicted values $\hat{y}$ and the ground-truth values $y$. $\sigma$ denotes the standard deviation and $\mu$ denotes the average of either the predicted or the ground-truth values.

\section{Handling Missing Modalities}
\label{section:missing_modalities}

We now describe another contribution of our paper, which is our approach to deal with missing modalities. 
In our case, a missing modality means that the modality is completely absent, thus the input sequence defined in Expression \ref{eq:inputSeq} is built with the remaining modalities.
By construction, our model will not break in the case a modality is missing. The attention mechanisms in our Transformer-based approach can accommodate missing modalities as explained below. In the case of the MMTE, if a modality is not present, the Tranformer encoder will simply attend to the remaining modalities. Similarly, in the AMMTD, the cross-modal attention will be able to attend to the remaining representations without the need to generate a replacement for the missing ones.

Even if our approach is capable to continue working in the case of missing modalities, its performance may be degraded. To increase the robustness of the model to missing modalities, we perform an \textit{optimized training} that we describe below.

First, we identify the most important modalities. To do this, we first train our model in a standard way, then test it without one modality at a time. We can then identify in which cases the performance is reduced more, meaning that the missing modalities in those cases should be important. Next, we retrain the model, without the important modalities a part of the time. Specifically, for each batch, we randomly select to eliminate the important modality $i$ with probability $\rho^i_\text{eliminate}$, and to keep all modalities with probability $\rho_\text{none} = 1 - \sum_{i=1}^{n} \rho^i_\text{eliminate}$, where $n$ is the number of important modalities.

Our reasoning behind this training strategy is that by hiding the important modalities, the model is forced to learn from the remaining ones, thus making the model more robust when those important modalities are missing. Moreover, we believe that this training strategy should lead to better results in general, even without missing modalities, as more information will be incorporated from all the modalities, rather than just relying on the important ones.

\section{Experimental Setup}
\label{section:setup}

We test our model on the task of recognizing time-continuous values of arousal and valence. In this section, we describe the dataset, features, and parameters of our model for these experiments.

\subsection{Dataset}
To evaluate our model, we use the Ulm-Trier Social Stress Test dataset (Ulm-TSST), which was presented for the Muse 2021 Challenge \cite{stappenMuSe2021Multimodal2021, stappenMuSe2021Challenge2021}.  This is a multimodal dataset, where participants were recorded in a stressful situation emulating a job interview, following the TSST protocol \cite{kirschbaumTrierSocialStress1993}. Each participant gave a 5 minutes speech supervised by two interviewers, who did not intervene during that time. Besides audio and video, the following physiological signals are collected: Electrocardiogram (ECG), Respiration (RESP), and heart rate (BPM). A transcription of the speech is also provided.

The data set was annotated by 3 raters giving continuous values of arousal and valence in the range $[-1, 1]$. The annotations are done in a time-continuous fashion every 0.5s. To aggregate the valence annotations from the 3 raters, Rater Aligned Annotation Weighting (RAAW) \cite{stappenMuSeToolboxMultimodalSentiment2021} is used. For arousal, the annotations corresponding to the lowest inter-rater agreement are discarded, and replaced by the subject's Electrodermal Activity (EDA) signal recorded during the session. The authors of the dataset do this because EDA has been demonstrated to be a good indicator of arousal \cite{bairdPhysiologicallyAdaptedGoldStandard2021}. Like it was done for valence, RAAW is used to aggregate the annotations from the two remaining raters and the EDA signal.

The dataset includes 69 samples, each being a 5-minute presentation given by a subject. In the original dataset, 41 samples are used as train set, 14 as validation set, and 14 as test set. Since annotations are not provided for the test set, we randomly pick 4 samples from the validation set and 6 from the train set to form a new test set with 10 samples. In summary, we have 35 samples in the train set, 10 in the validation set, and 10 in the test set. 

\subsection{Input Features}
We use audio, video, and physiological signals as input modalities, with features directly provided in Ulm-TSST. All features are aligned with annotations; that is, they are sampled at a rate of 2 Hz. For audio, we use extended Geneva Minimalistic Acoustic Parameter Set (eGeMAPS) features. For video, we use Facial Action Units (FAU) intensity. For physiological signals, the features are the concatenation of the values of ECG, RESP, and BPM. We selected these features by running some experiments in the baseline model provided by the authors of the Ulm-TSST dataset and selecting the features that lead to good performance.

\subsection{Model Hyperparameters and Training}

We optimize the hyperparameters of our model using the Ray Tune Framework \cite{liawTuneResearchPlatform2018} based on the validation set. Our model is parameterized as follows: we use a TCN with 6 layers and a kernel of size 9, with ReLU activation function. We have a model dimension of $d_\text{model} = 64$. In the Transformer encoder and the TDL, we use the GELU activation function. The size of the FFN inside the Transformers is $d_\text{model} \times 4 = 256$. We use a Transformer encoder with 2 attention heads and 2 layers. Our decoder is composed of a single TDL with one head. The FC layer in the ERN has a single hidden layer of $d_\text{model}/2 = 32$, with ReLU activation function. The bidirectional attention mask for the Transformer encoder has a \textit{mask\_length} of 50 seconds (100 time-steps).

During training, we segment each 5-minute sample into smaller samples, as suggested by Christ et al. \cite{christMuSe2022Multimodal2022a}. Searching across different options, we found that segments of 125 seconds (250 time-steps) with a hop size of 25 seconds (50 time-steps) work well in our experimental protocol.

We train our model with a batch size of 64 for a maximum of 100 epochs. We start with a learning rate of 0.0001, and halve it if the metric does not improve for 5 epochs on the validation set, and we early-stop the training if there is not improvement for 15 epochs. We use Adam optimizer with $B_1=0.9$ and $B_2=0.999$. We use a dropout rate of 0.2 throughout all the model.

\section{Experiments} 
\label{section:experiments}
This section presents and discusses the experimental results. For each experiment, we obtain 30 results by training the model with 30 different initialization seeds, reporting the average of those results.
We use the Holm-Bonferroni method to assert statistically significant difference in the comparisons in Section \ref{section:allModalities}.
For other results, we do a t-test using a threshold of p-value $<$ 0.05 to assert statistical significance, using a one-sided t-test to state that a result is significantly better than another, and a two-sided t-test to check for statistical difference.
We use as metric the Root-Mean-Square Error (RMSE) and the Concordance Correlation Coefficient (CCC) \cite{linConcordanceCorrelationCoefficient1989} between ground truths and predicted values.

\subsection{Performance with all Modalities Present} \label{section:allModalities}

\begin{table*}[t]
\centering
\caption{Comparison of our results with the baselines. The best result is in bold, the second best is underlined. The standard deviation is in parentheses. ($\bullet$), (\dag), (\ddag) indicate that our results are statistically significantly different than the late-fusion, MMTE+FC, and MMTE+LSTM baselines, respectively. (\textdownarrow) and (\textuparrow) indicate that a lower and a higher score is desirable respectively.}
\resizebox{\textwidth}{!}{
\begin{tabular}{c l l l l l}
\toprule[1pt]
& & \multicolumn{2}{c}{\textbf{Arousal}} & \multicolumn{2}{c}{\textbf{Valence}} \\ 
\textbf{\No} & \multicolumn{1}{c}{\textbf{Approach}} & \multicolumn{1}{c}{\textbf{RMSE}\textdownarrow} & \multicolumn{1}{c}{\textbf{CCC}\textuparrow} & \multicolumn{1}{c}{\textbf{RMSE}\textdownarrow} & \multicolumn{1}{c}{\textbf{CCC}\textuparrow}\\
\midrule[1pt]
1 & Late-fusion (LSTM) \cite{christMuSe2022Multimodal2022a}  & 0.3046 (0.0199) & 0.2702 (0.0258) & \textbf{0.1585} (0.0156) & 0.1273 (0.0528)\\
\midrule[0.5pt]
2& MMTE+FC (ERN)  & 0.3238 (0.0227) & 0.1388 (0.0682) & 0.1850 (0.0167) & 0.1221 (0.0309)\\
3& MMTE+LSTM      & 0.3189 (0.0244) & 0.1387 (0.0575) & 0.1842 (0.0653) & 0.0435 (0.0640)\\
\midrule[0.5pt]
4& MMTE+AMMTD (ours) & \underline{0.2948}\dag\ddag\ (0.0125) & \underline{0.3578}$\bullet$\dag\ddag\ (0.0317) & 0.1796 (0.0114) & \underline{0.1502}\dag\ddag\ (0.0272) \\
5& MMTE+AMMTD, optimized train (ours) & \textbf{0.2869$\bullet$\dag\ddag} (0.0120) & \textbf{0.3703$\bullet$\dag\ddag} (0.0351)  & \underline{0.1739}\ (0.0089) & \textbf{0.1656$\bullet$\dag\ddag} (0.0169) \\
\bottomrule[1pt]
\end{tabular}
}
\label{table:allModalities}
\end{table*}

We present in Table \ref{table:allModalities} the performance of our model, along with baseline approaches, when all modalities are present. Approach \No 1 corresponds to the baseline model developed for the Muse 2022 Challenge \cite{christMuSe2022Multimodal2022a}, where the Ulm-TSST dataset was presented. We use the provided code\footnote{https://github.com/EIHW/MuSe2022} and the original hyperparameters to evaluate this model with the same features we employ, using our partition of the Ulm-TSST dataset. This approach is based on Long Short-Term Memory (LSTM) networks and uses late-fusion to aggregate the different modalities. Approach \No 2 corresponds to a model where instead of using our AMMTD to process the representations from the MMTE, it uses directly the ERN. Recall that the ERN is a FC network that performs regression of the emotion values. Similarly, approach \No 3 uses an LSTM to process the outputs of the MMTE. The LSTM has 4 layers, with a hidden dimension of 32. We used a grid search to tune this LSTM. For both approaches \No 2 and 3, the input is the concatenation of all modalities per time-step. The last two entries in Table \ref{table:allModalities} correspond to the approach presented in this paper. Approach \No 4 is our model trained in a standard way, i.e. with all modalities present during training. Approach \No 5 is our model trained with our optimized training strategy as presented in Sections \ref{section:missing_modalities} and \ref{section:exMissingModalities}, i.e. hiding some modalities during training.

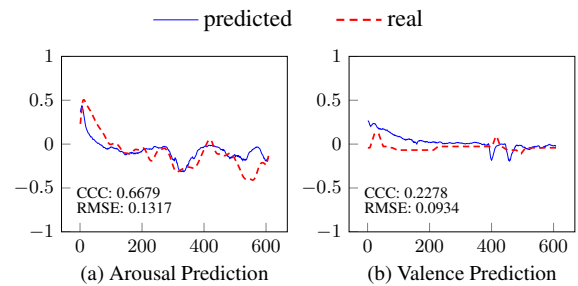
\begin{figure}[tb]
\pgfplotsset{
    width=0.3\textwidth,
    height=4.5cm,
    xtick pos=left,
    ytick pos=left,
    ymin=-1,
    ymax=1,
    ticklabel style = {font=\footnotesize},
}
\centering
\ref{named}\\
\begin{tikzpicture}[scale=0.8]
\begin{axis}[
    legend columns=-1,
    legend entries={predicted, real},
    legend to name=named,
    legend style={
        font=\footnotesize, draw=none,
        /tikz/every even column/.append style={column sep=0.5cm},
    },
    xlabel={(a) Arousal Prediction},
    ]
    \addplot +[mark=none, ultra thin] table[col sep=comma, x expr=\coordindex+1, y="value"] {arousal_68.csv};
    \addplot +[mark=none, thick, densely dashed] table[col sep=comma, x expr=\coordindex+1, y="ground_truth"] {arousal_68.csv};
\end{axis}
\node[right] at (0.1,0.65) {\tiny CCC: 0.6679};
\node[right] at (0.1,0.4) {\tiny RMSE: 0.1317};
\end{tikzpicture}
\begin{tikzpicture}[scale=0.8]
\begin{axis}[
    xlabel={(b) Valence Prediction},
    ]
    \addplot +[mark=none, ultra thin] table[col sep=comma, x expr=\coordindex+1, y="value"] {valence_68.csv};
    \addplot +[mark=none, thick, densely dashed] table[col sep=comma, x expr=\coordindex+1, y="ground_truth"] {valence_68.csv};
\end{axis}
\node[right] at (0.1,0.65) {\tiny CCC: 0.2278};
\node[right] at (0.1,0.4) {\tiny RMSE: 0.0934};
\end{tikzpicture}
\caption{Example of an output of our model with optimized training compared with the ground-truth, when predicting arousal (a) and valence (b) for the same sample.}
\label{fig:gtPredsExample}
\end{figure}

In Figure \ref{fig:gtPredsExample}, we present an example of the predictions of our model when using the optimized training strategy. For the same sample, we present the results when predicting arousal, Fig. \ref{fig:gtPredsExample}(a), and valence, Fig. \ref{fig:gtPredsExample}(b). As observed, the real valence values tend to be flat and have less variability than the arousal values, which we noted is a common occurrence in the dataset.

\subsubsection{Comparison to the LSTM-based baseline model} If we compare our approach with standard training (approach \No4 in Table \ref{table:allModalities}) with the LSTM baseline (approach \No 1), we can see that in all metrics except for valence RMSE, our model performs better than the LSTM baseline, demonstrating that in general, our Tranformer-based approach is well suited for this task.

\subsubsection{Comparison with other predictors} To test our idea of using cross-attention to weigh the input modalities and an auto-regressive approach to incorporate past predictions, we compare our approach with standard training (approach \No 4), with approaches \No 2 and \No 3 in Table \ref{table:allModalities}, that use the ERN and an LSTM respectively instead of our AMMTD. The results show that our AMMTD module increases the performance in most of the metrics, demonstrating the effectiveness of our ideas of using cross-attention and auto-regression. The performance of our approach is statistically significantly better for all the metrics except for valence RSME, where although our approach outperforms both baselines, the improvement is not statistically significant.

In general, the baseline models perform well in terms of the RMSE metric when predicting valence, but our model performs better in terms of the CCC metric. We hypothesize that this behavior is produced because the simpler architecture of the baselines is good enough to predict flat sequences of valence values that are close enough to the flat ground-truth. On the other hand, those approaches fail to predict the small changes in the valence values, penalizing the CCC score.

\subsubsection{Comparison with our optimized training strategy}
The results presented in entries \No 4 and \No 5 in Table \ref{table:allModalities} show that our optimized training approach, designed to improve the handling of missing modalities, also has the desirable effect of increasing the performance of the model when all modalities are present. For example, arousal RMSE decreases from 0.2948 to 0.2869 and valence CCC increases from 0.1502 to 0.1656. As we expected, the model seems to learn to use more information from the \emph{weak} modalities, improving the overall performance.

\subsection{Accommodating Missing Modalities}\label{section:exMissingModalities}

\begin{table*}[t]
\centering
\caption{Summary of results when modalities are missing, for standard training and our optimized training strategy. We use bold font to indicate that the result is better than its counterpart trained in a different fashion, and if it is statistically significantly better we indicate this with the symbol (\ddag). We use a checkmark (\checkmark) to indicate that a result obtained with a modality missing is not statistically significantly different than the result obtained with all the modalities.
}
\resizebox{\textwidth}{!}{
\begin{tabular}{l l l l l l l l l l }
\toprule[1pt]
& & \multicolumn{2}{c}{\textbf{All Modalities}} & \multicolumn{2}{c}{\textbf{Missing Audio}} & \multicolumn{2}{c}{\textbf{Missing Video}} & \multicolumn{2}{c}{\textbf{Missing Physio}} \\ 
& \textbf{Training Mode} & \textbf{RMSE}\textdownarrow  & \textbf{CCC}\textuparrow & \textbf{RMSE}\textdownarrow  & \textbf{CCC}\textuparrow  & \textbf{RMSE}\textdownarrow  & \textbf{CCC}\textuparrow  & \textbf{RMSE}\textdownarrow  & \textbf{CCC}\textuparrow \\
\midrule[1pt]
\multirow{2}{*}{\textbf{AROUSAL}} & Standard & 0.2948 & 0.3578 & 0.2926\checkmark & 0.3589\checkmark & 0.3252 & 0.2713 & 0.2920\checkmark & 0.3539\checkmark \\
                                  & Optimized & \textbf{0.2869}\ddag & \textbf{0.3703} & \textbf{0.2850}\checkmark\ddag & \textbf{0.3644}\checkmark & \textbf{0.3249} & \textbf{0.2984}\ddag  & \textbf{0.2878}\checkmark & \textbf{0.3571}\checkmark \\
\midrule[0.5pt]
\multirow{2}{*}{\textbf{VALENCE}} & Standard & 0.1796 & 0.1502 & 0.2533 & 0.0738 & 0.2170 & 0.1564\checkmark & 0.1808\checkmark & 0.1486\checkmark \\
& Optimized & \textbf{0.1739}\ddag & \textbf{0.1656}\ddag & \textbf{0.2052}\ddag & \textbf{0.1170}\ddag & \textbf{0.1809}\checkmark\ddag & \textbf{0.1676}\checkmark\ddag  & \textbf{0.1746}\checkmark\ddag & \textbf{0.1637}\checkmark\ddag \\
\bottomrule[1pt]
\end{tabular}
}
\label{table:missingModalities}
\end{table*}

We present in Table \ref{table:missingModalities} the results of experiments we conducted when a modality is missing.

First, we analyze the case where we are predicting arousal with the model trained in a standard way. In Table \ref{table:missingModalities}, we see that there is no significant performance degradation when the audio or physiological modalities are missing. For example, when audio is missing, RMSE goes from 0.2948 to 0.2926, and CCC goes from 0.3578 to 0.3589. These differences are not statistically significant, as indicated by the checkmark (\checkmark). We also see that performance drops significantly when the video modality is missing, with RMSE increasing from 0.2948 to 0.3252 and CCC decreasing from 0.3578 to 0.2713. These results confirm that our model continues to accurately predict arousal when a modality is missing, although performance is reduced in some cases. 

For valence, we see that there is no significant performance degradation when physiological signals are missing, with RMSE going from 0.1796 to 0.1808 and CCC going from 0.1502 to 0.1486. On the contrary, the model performance drops significantly when the audio or video modalities are missing. For example, RMSE increases from 0.1796 to 0.2533 when the audio modality is missing and increases to 0.2170 when the video modality is missing. Much like for arousal, these results show that our model continues to estimate valence when a modality is missing, albeit with a drop in performance.

With these results, we can identify the most important modalities for our model, by identifying the modalities that when missing, induce a significant drop in performance. When predicting arousal, the most important modality is video, and for valence the most important modalities are audio and video. 

Models tend to rely heavily on the important modalities, even though other modalities may carry useful features for recognizing emotions. To improve this, we apply our optimized training strategy of hiding important modalities during parts of the training, forcing the model to rely on other modalities. Specifically, we use the strategy described in Section \ref{section:missing_modalities}, eliminating the video modality with probability $\rho^\text{video}_\text{eliminate} = 0.25$, and maintaining all modalities with probability $\rho_\text{none} = 0.75$ when training for arousal prediction. For valence, we use $\rho^\text{audio}_\text{eliminate} = 0.333$, $\rho^\text{video}_\text{eliminate} = 0.333$ and $\rho_\text{none} = 0.334$. These probabilities were found empirically by testing several configurations and keeping the best ones for the validation set.

We see in Table \ref{table:missingModalities} that our optimized training strategy improves all results when any modality is missing. For example, when the physiological signals are missing, CCC improves from 0.3539 to 0.3571 when predicting arousal and from 0.1486 to 0.1637 when predicting valence. Notably, the improvement is statistically significant, as indicated by the double dagger (\ddag), in all cases when the \emph{important} modalities are missing, except for RMSE when predicting arousal with the video modality missing. In addition, the performance of the model improves even when all modalities are present. For instance, RMSE decreases from 0.2948 to 2869 for arousal and from 0.1796 to 0.1739 for valence.

This shows that our optimized training strategy works well, making our model less reliant on the important modalities and using more information from the other ones. This strategy improves the performance of our approach not only when a modality is missing, but also when all modalities are present.

\section{Conclusions and Perspectives}
\label{section:conclusion}
 
In this work, we presented a novel Transformer-based architecture with the coupling of self- and cross-attention mechanisms for emotion recognition from multimodal signals. 
We experimentally showed, using the Ulm-TSST dataset, that our proposal can competitively recognize emotional valence and arousal. 
In addition, we demonstrated that our optimized learning strategy improves performance. 
Consequently, our architecture is capable of reaching high performances for emotion recognition even with missing modalities.

Future works include investigating new ways to pre-train multimodal models with contrastive learning strategies. Metric learning between different signals, sharing the same semantic meaning, could positively influence the whole system to build strong correlations between input sequences produced at different moments in time. Moreover, learning similarities between signals even when some modalities are missing may help to better understand how pretraining on Transformer-based models behave for multimodal emotion recognition.

\section*{Ethical Impact Statement}

We consider two main issues: privacy and harmful applications. First, regarding privacy, knowing the emotional state of a person leads to privacy concerns since it is evident that the emotional state is a private matter. Therefore, privacy should be guaranteed. There might be cases that people involved might agree to share this private information, as in the case of people participating in the dataset that we use. That is why we adhere to the dataset usage agreement. In the general case, to avoid privacy concerns, we think the emotional record of a person should be treated at the same level of privacy that medical records are treated.

The second issue has to do with potentially harmful applications. Although in this work we do not develop an application that uses the emotional state of a person as input, we are aware that bad actors may use emotion recognition techniques in unethical ways. For example, knowing the emotional state of a person can be used to manipulate their behavior. Direct control of this is out of our hands, and therefore, we think the affective computing community should pressure governmental entities for laws and ways to control these types of applications.

\vspace{0.25cm}
{\small
\textbf{Acknowledgements:} This work has been partially supported by the MIAI Multidisciplinary AI Institute at the Univ. Grenoble Alpes:  (MIAI@Grenoble Alpes - ANR-19-P3IA-0003).}

\bibliographystyle{IEEEbib}
\bibliography{refs.bib}

\end{document}